\pdfoutput=1

\documentclass[11pt]{article}

\usepackage{algorithm}
\usepackage{algpseudocode}
\usepackage{amsmath}
\usepackage{multirow} 
\usepackage{float}
\usepackage{enumitem}
\usepackage[final]{acl}

\usepackage{times}
\usepackage{latexsym}

\usepackage[T1]{fontenc}

\usepackage[utf8]{inputenc}

\usepackage{microtype}

\usepackage{inconsolata}

\usepackage{graphicx}

\title{LoRA-PAR: A Flexible Dual-System LoRA Partitioning Approach to Efficient LLM Fine-Tuning}

\author{
 \textbf{Yining Huang\textsuperscript{1}}\thanks{\href{mailto:huangyining1987@gmail.com}{huangyining1987@gmail.com}},
 \textbf{Bin Li\textsuperscript{2}}\thanks{\href{mailto:b.li2@siat.ac.cn}{b.li2@siat.ac.cn}},
 \textbf{Keke Tang\textsuperscript{3,4}}\thanks{\href{mailto:tkk2012@gmail.com}{tkk2012@gmail.com}},
 \textbf{Meilian Chen}\thanks{\href{mailto:523062863@qq.com}{523062863@qq.com}}
\\
 \textsuperscript{1}School of Politics and Public Administration, South China Normal University
\\
 \textsuperscript{2}Shenzhen Institute of Advanced Technology, Chinese Academy of Sciences
\\
 \textsuperscript{3}University of Chinese Academy of Sciences
\\
 \textsuperscript{4}Shenyang institute of computing technology, Chinese academy of sciences
\\
 \small{
 \textbf{Correspondence:} 
 \href{mailto:huangyining1987@gmail.com}{huangyining1987@gmail.com}, 
 \href{b.li2@siat.ac.cn}{b.li2@siat.ac.cn}
 }
}

\begin{document}
\maketitle
\begin{abstract}
    Large-scale generative models like DeepSeek-R1 and OpenAI-O1 benefit substantially from chain-of-thought (CoT) reasoning, yet pushing their performance typically requires vast data, large model sizes, and full-parameter fine-tuning. While parameter‐efficient fine‐tuning (PEFT) helps reduce cost, most existing approaches primarily address domain adaptation or layer‐wise allocation rather than explicitly tailoring data and parameters to different response demands. Inspired by “Thinking, Fast and Slow,” which characterizes two distinct modes of thought—System 1 (fast, intuitive, often automatic) and System 2 (slower, more deliberative and analytic)—we draw an analogy that different “subregions” of an LLM’s parameters might similarly specialize for tasks that demand quick, intuitive responses versus those requiring multi-step logical reasoning. Therefore, we propose \textbf{LoRA-PAR}, a dual-system LoRA framework that partitions both data and parameters by System 1 or System 2 demands, using fewer yet more focused parameters for each task. Specifically, we classify task data via multi-model role-playing and voting, and partition parameters based on importance scoring, then adopt a two-stage fine-tuning strategy of training System 1 tasks with supervised fine-tuning (SFT) to enhance knowledge and intuition and refine System 2 tasks with reinforcement learning (RL) to reinforce deeper logical deliberation next. Extensive experiments show that the two-stage fine-tuning strategy, SFT and RL, lowers active parameter usage while matching or surpassing SOTA PEFT baselines.

\end{abstract}

\begin{figure}[t]
  \includegraphics[width=\columnwidth]{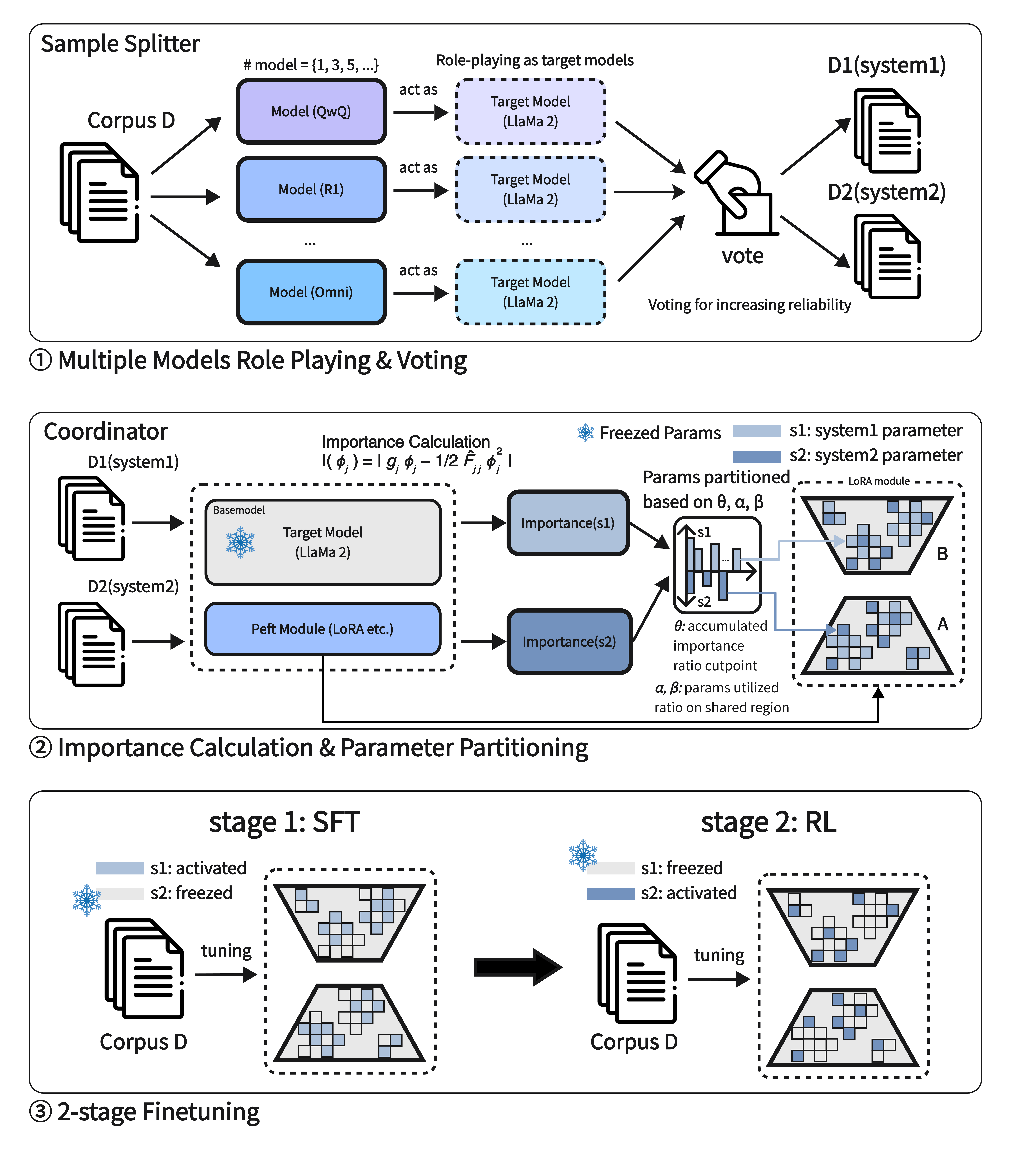}
  \caption{Overview of our proposed workflow. (1) Sample Splitter: Multiple LLMs role-play the target model and vote to classify the corpus into System 1 and System 2 data. (2) Coordinator: We calculate importance scores for each LoRA parameter in both tasks and selectively partition them based on cumulative ranking. (3) A two‐stage fine‐tuning pipeline (SFT then RL) trains the shared or task‐specific parameters, ensuring efficiency while preserving both language fluency and logical reasoning performance.}
  \label{fig:workflow}
  \vspace{-0.2cm}
\end{figure}

\begin{figure*}[t]
  \includegraphics[width=1\linewidth]{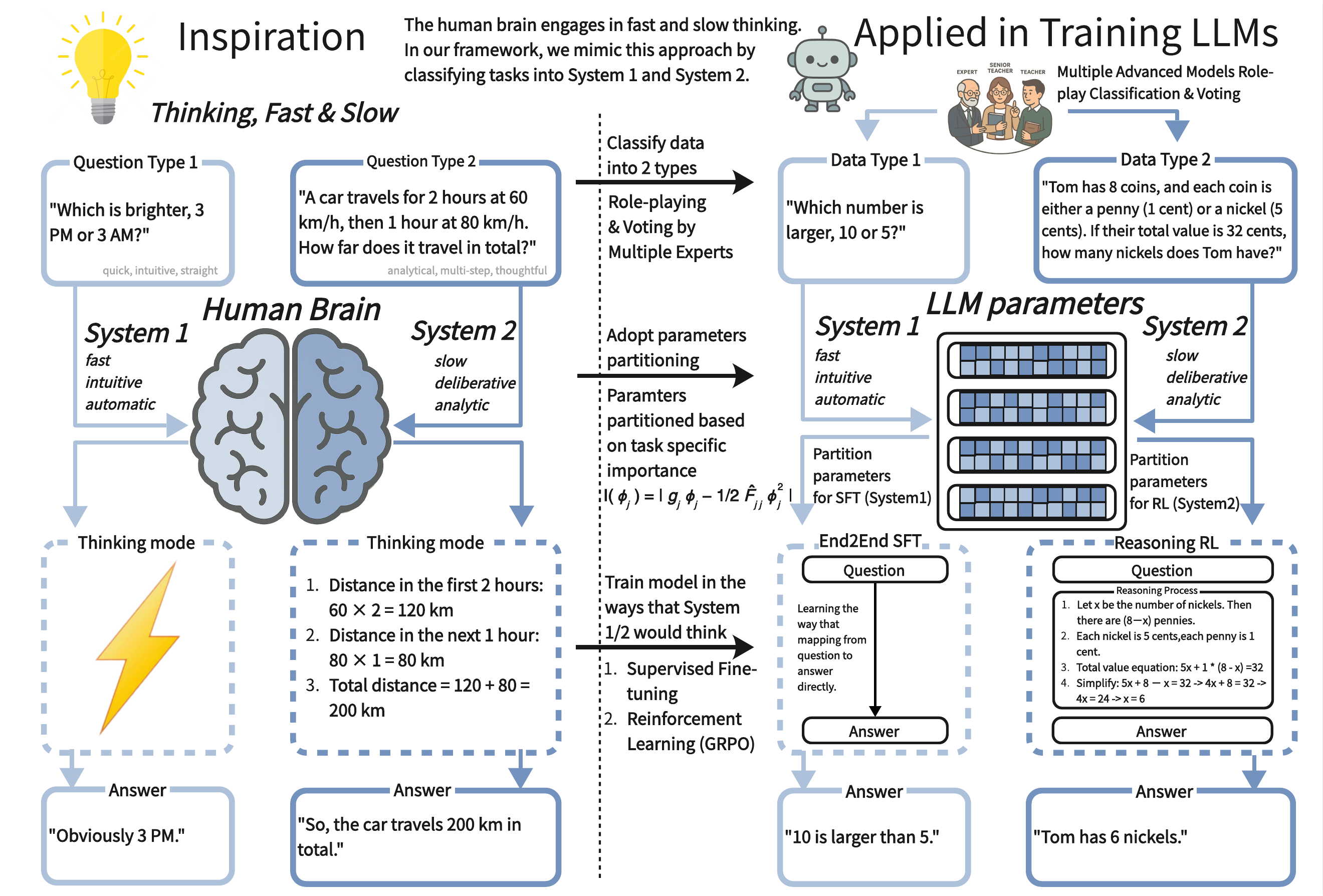} \hfill
    \vspace{-0.3cm}
  \caption {Inspired by Thinking, Fast and Slow, we introduce System 1 and System 2 into LLM training. (1) We assume subregions of model parameters can mirror the brain’s distinct cognitive modes; (2) we classify data into System 1 or System 2 via expert role-play voting and allocate parameters accordingly, then parameters will be partitioned based on importance score with respect to different task data; and (3) we train System 1 with end-to-end SFT for direct mappings from question to answer, while System 2 uses RL (GRPO) to encourage multi-step reasoning.}
  \label{fig:main}
    \vspace{-0.3cm}
\end{figure*}

\section{Introduction}
Large language models (LLMs) such as DeepSeek-R1 \citep{DeepSeekAI2025DeepSeekR1IR} and OpenAI-O1 \citep{Contributors2024OpenAIOS} have shown remarkable progress in complex reasoning when equipped with CoT prompts. However, pushing their performance to new levels often relies on massive datasets and full-parameter fine-tuning, requiring considerable compute and large model sizes. To alleviate this burden, parameter-efficient fine-tuning (PEFT) approaches have emerged as a promising alternative. Most existing PEFT methods, however, predominantly insert uniform adapter modules (e.g., LoRA \citep{Hu2021LoRALA}) and do not specifically tailor their parameter configurations to the unique demands of different tasks or reasoning levels. Although there have been some recent attempts to design more task- or data-aware PEFT solutions \citep{Zhang2023AdaptiveBA, Zhang2023TowardsAP}, these works largely focus on domain adaptation or layer-wise parameter allocation rather than explicitly targeting more advanced, multi-step reasoning capabilities.
\par
Meanwhile, inspired by Thinking, Fast and Slow \citep{Kahneman2011ThinkingFA}, we incorporate the dual‐system concept into parameter-efficient fine-tuning for LLMs. Specifically, as shown in \autoref{fig:main}, we draw on the idea that the human brain engages partially distinct neural processes for System 1 versus System 2. 
Recent studies further evidence that large language models can manifest or benefit from distinct “fast” vs.\ “slow” modes: \citep{Hagendorff2022HumanlikeIB} exhibit human-like intuitive biases, \citep{Pan2024DynaThinkFO} propose dynamic decision mechanisms inspired by Kahneman’s framework, and more general discussions support bridging cognitive dual-process theories and AI \citep{Booch2020ThinkingFA}. 
By analogy, we posit that an LLM’s parameters can be partitioned into “subregions” specialized for different response demands. We implement this in three steps: (1) we use multi-expert role‐play and voting to classify each training instance into System 1 or System 2 tasks, ensuring quick, direct “fast thinking” problems are kept separate from more deliberative multi-step tasks; (2) we then allocate different subsets of parameters of LoRA modules (via importance-based partitioning) for System 1 and System 2, akin to activating distinct cognitive modes; (3) we train System 1 parameters with end-to-end SFT for direct question–answer mapping, and refine System 2 parameters using reinforcement learning (GRPO \citep{Shao2024DeepSeekMathPT}), similar to how models like DeepSeek‐R1 \citep{DeepSeekAI2025DeepSeekR1IR} achieve deeper chain‐of‐thought style reasoning. In this way, our approach remains within the lightweight scope of PEFT, while still capturing the dual‐process benefits of human cognition—fast, intuitive responses and methodical, step‐by‐step logic.

\section{Related Work}
\subsection{Parameter Importance Calculation and Pruning}
SparseGPT \citep{Frantar2023SparseGPTML} effectively prunes large-scale LLM parameters without retraining, drastically reducing model size with minimal performance loss. Wanda \citep{Sun2023ASA} employs activation-aware magnitude pruning without retraining, significantly outperforming traditional magnitude-based methods. LLM-Pruner \citep{Ma2023LLMPrunerOT} identifies and removes structurally redundant components via gradient-based scoring, retaining general multitask capabilities. Týr-the-Pruner \citep{Li2025TrthePrunerUA} applies second-order Taylor approximations for global structured pruning, achieving high sparsity levels with minimal accuracy loss.

\subsection{Selective Freezing and Dual-Stage Training}

LIMA \citep{Zhou2023LIMALI} shows minimal fine-tuning effectively aligns pretrained models, implying substantial portions of models can remain frozen without loss of knowledge. ILA \citep{Shi2024UnderstandingLS} develops an analysis technique to selectively freeze non-critical layers, improving fine-tuning efficiency and performance. Safety Layer Freezing \citep{Li2024SafetyLI} suggests freezing identified "safety-critical" layers during further fine-tuning to preserve original alignment and safety behaviors.

\subsection{LoRA and PEFT Variants}

LoRA \citep{Hu2021LoRALA} introduces low-rank adaptation, drastically reducing fine-tuning overhead by freezing most parameters while updating small adapter matrices. PiSSA \citep{Meng2024PiSSAPS} initializes LoRA adapters using pretrained singular vectors, accelerating convergence and boosting task accuracy. OLoRA \citep{Bykakyz2024OLoRAOL} enhances LoRA initialization with orthonormal matrices, significantly accelerating fine-tuning convergence. QLoRA \citep{Dettmers2023QLoRAEF} enables highly efficient 4-bit quantized fine-tuning of large models, drastically lowering computational requirements without performance loss. LoRA+ \citep{Hayou2024LoRAEL} optimizes LoRA fine-tuning via learning-rate scaling adjustments, achieving faster convergence and higher accuracy.

\subsection{Training \& Data Strategies for Better LLM Reasoning}

GSM-DC builds math problems as symbolic dependency graphs to inject controlled irrelevant context and adds stepwise evaluation to improve reasoning robustness \citep{yang2025llm}. Toolformer self-supervises API-call annotations to help LLMs learn when/how to use external tools, which improves reasoning on tasks needing computation or lookup \citep{schick2023toolformer}. AlphaGeometry synthesizes millions of theorems and proofs to train a neuro-symbolic system that reaches Olympiad-level geometry performance \citep{trinh2024solving}. A NeurIPS'24 study proposes a neuro-symbolic data-generation pipeline that mutates math problems and verifies them with solvers, then realigns LLMs on the generated set to surpass state-of-the-art baselines \citep{li2024neuro}. 

\section{Method}
\label{sec:Method}
\subsection{Overall Workflow}
The overall workflow of our proposal proceeds as follows. First, multiple teacher LLMs vote to label each query as fast, single‐step (System 1) or multi‐step reasoning (System 2). Next, we compute parameter importance in LoRA and keep only the most cumulative‐importance score parameters for each system, identifying a shared subset that are important to both. Finally, we apply a two‐stage fine‐tuning strategy, using SFT for System 1 tasks and RL for System 2. Shared parameters can be partially activated in both stages, controlled by $\alpha$ and $\beta$. This design efficiently addresses “fast vs.\ slow thinking” within a single LLM by freezing irrelevant parameters and focusing updates on the most crucial subregions.

\subsection{Multi-Model Role‐Play and Voting for Data Classification}
\label{sec:data_classification}
\begin{figure}[t]
  \includegraphics[width=\columnwidth]{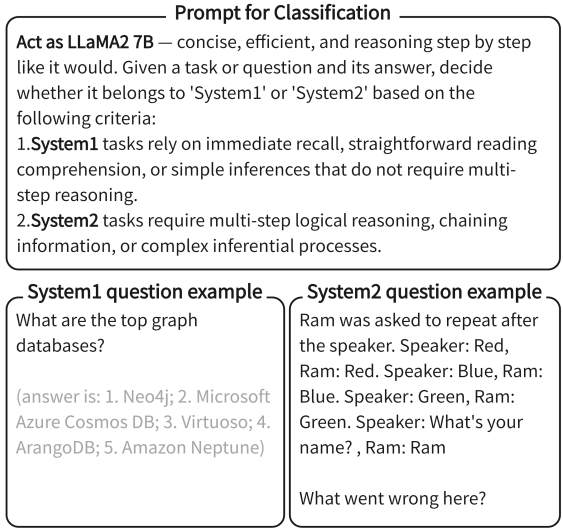}
  \caption{The prompt for classification and example questions of System 1 \& System 2}
  \label{fig:prompt_example}
      \vspace{-0.2cm}
\end{figure}

Before partitioning model parameters into different mode of thinking, questions are needed to be identified to fall into which category. Rather than relying on a single classifier—which may be error‐prone or biased—we design a multi‐model role‐playing approach. Here, several advanced LLMs (like the “teachers”) each act to be the “target” model (like the “student”) and classify the questions accordingly. Because these teacher models typically have broader pretraining coverage, they can approximate how the student would perceive the question type—either System 1 or System 2. The prompt of role-playing and example questions of System 1 and System 2 are shown in \autoref{fig:prompt_example}.

As shown in the upper panel of \autoref{fig:workflow} (cf. “Sample Splitter”), each teacher independently provides a classification, and we then apply a voting procedure to aggregate these judgments. This ensures that disagreements—arising from the teachers’ differing architectures or training histories—are resolved in a robust manner. The resulting labeled subsets, $D_1$ (System 1) and $D_2$ (System 2), feed into the subsequent modules, where they guide parameter partitioning and two‐stage training.

\subsection{Parameter Importance Calculation for Subregion Partitioning}
\label{sec:partitioning}
After classifying questions, the next step is to determine which LoRA parameters should be “activated” for each category. We adopt LoRA rather than full‐parameter fine‐tuning to preserve the base model’s global knowledge and enable a modular strategy of activate or freeze for System 1 and System 2 tasks. 
The partitioning process parallels how different regions of the human brain are activated in response to different cognitive demands \citep{Kahneman2011ThinkingFA}. In large language models, parameter gradients serve as an analogue to neural activations. If the gradient for a certain parameter is large, it implies that parameter is critical in correcting output errors for a particular task.
To improve the model’s ability to answer different types of questions (System 1 or System 2), we apply a mask to ignore the prompt and context tokens in the loss computation—i.e., we focus only on output positions. This ensures our importance scores emphasize each parameter’s contribution to generating the correct final answer rather than merely modeling the prompt text.

\paragraph{Computing Importance Scores.}
In practice, we attach LoRA modules at positions of Q/K/V/Gate/Up/Down within the target model layers. Let $\phi_j$ denote an individual LoRA parameter. We measure its importance via a Taylor expansion of the masked cross‐entropy loss $L(\cdot)$ up to second order:
\begin{align}
\Delta L(\phi_j) &\approx
\left| g_j \,\phi_j - \tfrac{1}{2}\,\hat{F}_{jj}\,\phi_j^2 \right| \label{eq:deltaL} \\
g_j &= \frac{\partial L}{\partial \phi_j}, \quad
\hat{F}_{jj} \approx \frac{1}{N} \sum_{k=1}^N
\left( \frac{\partial L_k}{\partial \phi_j} \right)^2 \label{eq:grad_F} \\
I(\phi_j) &= \left| g_j \,\phi_j - \tfrac{1}{2}\,\hat{F}_{jj}\,\phi_j^2 \right| \label{eq:importance}
\end{align}
Here, $g_j$is the gradient of the masked loss w.r.t.\\$\phi_j$, and $\hat{F}_{jj}$ is the diagonal of the Fisher matrix approximated from per‐example gradients $L_k$. Focusing on the \emph{output tokens} aligns parameter importance with the model’s ability to produce correct answers.

\paragraph{Selecting and Freezing Parameters.}
We rank $\{\phi_j\}$ by $I(\phi_j)$ and choose the top fraction (controlled by $\theta$) as the “activated” subregion for each System. During training, activated parameters remain learnable while the rest are frozen, reducing overhead. Some parameters may appear crucial for both System 1 and System 2; these “overlapping” parameters are shared across the two fine‐tuning stages. By partitioning parameters in this manner, our framework moves closer to the neural analogy that different “subregions” are engaged for tasks that demand fast vs.\ slow thinking.

\subsection{Two-Stage Fine-tuning Strategy with Importance-Based Parameter Selection}
\label{sec:two_stage}
Building on the importance scores computed in \S\ref{sec:partitioning}, we now formalize how to (i) determine how many parameters to activate for each system, (ii) handle the overlap between System 1 and System 2 parameters, and (iii) schedule the fine‐tuning process in two distinct stages. As shown in Algorithm~\autoref{alg:train}, our approach hinges on three hyperparameters—$\theta$, $\alpha$ and $\beta$—that control which parameters and how many are updated for System 1 (SFT) and System 2 (RL).

\paragraph{Threshold $\theta$: Selecting the Most Important Parameters.}
From the parameter‐importance visualization (see Figure~\ref{fig:weight_importance_plots}), we observe that System 1 and System 2 each rely on a partially disjoint set of LoRA parameters, with a notable overlap. Moreover, each dataset contains many “low‐impact” parameters whose importance is near zero for both systems. We introduce a cumulative‐importance threshold $\theta$. Specifically, for each system’s importance ranking, we keep only the most important subset of parameters with cumulative‐importance score over $\theta$, discarding the tail of negligible‐importance parameters to reduce overhead and avoid unnecessary updates. For instance, setting $\theta=0.9$ means we retain only the parameters with cumulative‐importance score over 90\%, they are crucial for System 1 and System 2 tasks, respectively.\footnote{The $\theta$ is not the ratio of parameters used in the following training procedure. In the situation of puting LoRA on QKVGUD of LLaMa2 7B, if we set $\theta=0.9$, there are approximately 40\% of parameters that will treated as important to System 1/2.} 

\paragraph{Activation Fractions $\alpha$ and $\beta$: Handling Overlap.}
Applying $\theta$ individually to System\,1 and System\,2 yields two top‐ranked sets of LoRA parameters, which partially overlap. Concretely, some parameters rank among the top‐ranked for \emph{both} systems; we call these ``shared'' (see the purple region in Figure~\ref{fig:weight_importance_plots}). We thus introduce two activation fractions, $\alpha$ and $\beta$, to control how many of these shared parameters are updated during the two training stages:

\begin{itemize}
\item \textbf{Stage~1 (SFT on System\,1 tasks):} We activate (a) all parameters in the System\,1‐only subset and (b) an $\alpha$ fraction of the shared parameters. If $\alpha<1$, we only partially train the shared region in this stage.
\item \textbf{Stage~2 (RL on System\,2 tasks):} We then activate (a) all parameters in the System\,2‐only subset and (b) a $\beta$ fraction of the shared parameters. The rest remain frozen, enabling us to flexibly allocate more (or fewer) shared parameters to System\,2 based on $\beta$.
\end{itemize}

By varying $\alpha$ and $\beta$, we fine‐tune the balance between ``fast, direct'' adaptation for System\,1 and ``multi‐step, deliberative'' adaptation for System\,2, ensuring that parameters useful for both can be partially or fully trained in each stage as needed.

\paragraph{Why Two Distinct Stages (SFT then RL)?}
We adopt SFT $\longrightarrow$ RL following practices in OpenAI GPT, DeepSeek-R1, and related literature on multi‐stage language model training. System 1 tasks—quick, direct Q\&A—are naturally suited to end-to-end SFT, which establishes “fast‐thinking” capability without delving into complex reasoning. This “knowledge foundation” helps bootstrap the second stage, where RL encourages step‐by‐step logical reasoning for System 2 tasks (akin to a “slow‐thinking” process). In essence, RL refines and extends the capabilities acquired via SFT, rewarding correct multi‐step strategies rather than just direct answers.

\paragraph{Putting It All Together.}
Algorithm~\ref{alg:train} outlines these steps more formally. In Stage 1 (SFT), only the System 1-only subset plus $\alpha$-portion of shared parameters are trained; in Stage 2 (RL), only the System 2-only subset plus $\beta$-portion of shared parameters are updated. This design ensures each system’s specialized subregion is honed for its respective tasks, while shared parameters can flexibly contribute to both fast and slow thinking modes. 

\section{Experiment}
\subsection{Experimental Setup.} 
We begin by partitioning each dataset via multi‐model role‐playing and voting \S\ref{sec:data_classification}, then compute LoRA parameter importance and keep the top‐ranked for each system \S\ref{sec:partitioning}. Training proceeds in two stages \S\ref{sec:two_stage}: (1) SFT for System 1, and (2) RL for System 2, with shared parameters managed by $\alpha$ and $\beta$. We measure accuracy across GSM8K \citep{cobbe2021gsm8k}, MMLU \citep{hendryckstest2021} (trained using Dolly15K \citep{DatabricksBlog2023DollyV2} or OpenPlatypus \citep{platypus2023}), and HumanEval \citep{chen2021evaluating} (code tasks), comparing our approach to LoRA \citep{Hu2021LoRALA}, OLoRA \citep{Bykakyz2024OLoRAOL}, PiSSA \citep{Meng2024PiSSAPS}, and PiSSA+RL, all based on LLaMA2 7B. Key hyperparameters include $\theta$ (fraction of top‐ranked parameters), $\alpha$, $\beta$ (activation fractions for overlapping parameters), and 1–2 training epochs for each baseline.

\begin{figure*}[t]
  \includegraphics[width=1\linewidth]{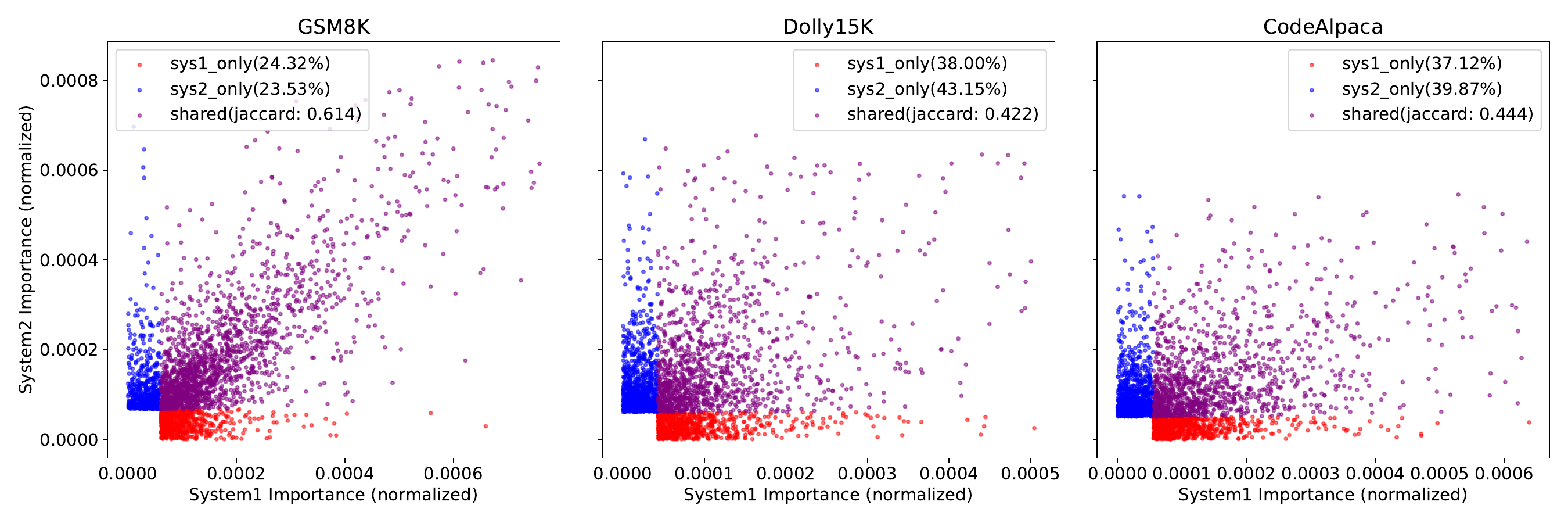} \hfill
  \vspace{-0.7cm}
  \caption {Scatter plots of LoRA parameter importance for System 1 (x‐axis) vs.\ System 2 (y‐axis) with a cumulative importance cutpoint $\theta=0.9$. Each dot is then labeled as System 1 only (red), System 2 only (blue), or shared (purple), the legend indicate the fraction of non‐overlapping parameters and the Jaccard overlap between System 1 and System 2 top sets. Notice that a substantial portion of parameters specialize in one system while a moderate number are shared, highlighting the potential for distinct subregions of LoRA parameters in multi‐stage fine‐tuning.}
  \label{fig:weight_importance_plots}
  \vspace{-0.4cm}
\end{figure*}

\begin{table}[t]
  \centering
  \begin{tabular}{lc}
    \hline
    \textbf{Data Classification Strategy} & \textbf{Performance} \\
    \hline
    QwQ w/o role play         & 25.32 \\
    QwQ w/ role play          & 26.23 \\
    Deepseek-R1 w/ role play           & 26.84 \\
    Random partition                            & 25.85 \\
    Role play + voting (n=3)    & 27.07\\
    Role play + voting (n=5)    & \textbf{27.60} \\
    \hline
  \end{tabular}
  \caption{
  \textbf{Performance on GSM8K under different data classification approaches.}
All runs use LLaMA2 7B + LoRA (QKV/Gate/Up/Down) and SFT. ``n'' indicates the number of external LLM used to vote.}
  \label{tab:rp_vo}
\vspace{-0.5cm}
\end{table}

\subsection{Role‐Playing and Voting for Data Classification}
We first verify the role‐playing and voting approach introduced in \S\ref{sec:data_classification} by comparing various data classification strategies on GSM8K. Specifically, we contrast (a) a single model without role‐play, (b) a single model prompted to “act as” LLaMA2 7B, (c) random partitioning, and (d) multiple models with role‐play plus voting. As shown in Table \ref{tab:rp_vo}, the multi‐model role‐play+voting setup achieves the highest performance. Prompting an external LLM to imitate the target model’s decision boundary (role‐playing) reduces misclassification compared to its default inference style, while the voting ensemble mitigates individual biases and yields more robust splits. This result aligns with our intuition that combining multiple “teacher” perspectives better approximates how LLaMA2 7B would distinguish System 1 vs.\ System 2 questions, ultimately enhancing downstream fine‐tuning.

\vspace{-0.3cm}
\subsection{Adaptive Parameter Usage via \texorpdfstring{$\theta$}{theta}}
\label{sec:adaptive_theta}

We next investigate how varying the cumulative importance cutpoint $\theta$ (from 0\% to 100\%) affects both the number of LoRA parameters activated and the resulting performance under SFT. In essence, $\theta$ dictates \emph{which} and \emph{how many} parameters are updated, as introduced in \S\ref{sec:two_stage}. For each setting, we compare three LoRA module configurations—QKV, GUD, and QKVGUD—against a random‐selection baseline that picks the same fraction of parameters but without regard to importance.

\begin{algorithm}[t]
\caption{Two-stage Fine-tuning} 
\label{alg:train}
\begin{algorithmic}[1]
\State \textbf{Input:} LoRA parameters $\{\phi_i\}_{i=1}^N$, importance scores $s_1(i)$, $s_2(i)$, thresholds $\theta, \alpha, \beta$
\State \textbf{Step 1 (Partition):} 
\begin{indent}
   \State Sort parameters by $s_1(i)$/$s_2(i)$ and keep the top important fraction as $S_1$/$S_2$.
   \State $\Omega_1\text{-only}=S_1 \setminus S_2,\; \Omega_2\text{-only} = S_2 \setminus S_1,\; \Omega_{\text{shared}} = S_1 \cap S_2$.
\end{indent}
\State \textbf{Step 2 (Stage 1: SFT):}
\begin{indent}
   \State Activate all parameters in $\Omega_1\text{-only}$.
   \State For each $\phi_i \in \Omega_{\text{shared}}$, activate it if it is in the top-$\alpha$ fraction by $s_1(i)$.
   \State Freeze all other parameters.
\end{indent}
\State \textbf{Step 3 (Stage 2: RL):}
\begin{indent}
   \State Activate all parameters in $\Omega_2\text{-only}$.
   \State For each $\phi_i \in \Omega_{\text{shared}}$, activate it if it is in the top-$\beta$ fraction by $s_2(i)$.
   \State Freeze all other parameters.
\end{indent}
\end{algorithmic}

\end{algorithm}

\begin{figure*}[t]
  \includegraphics[width=1\linewidth]{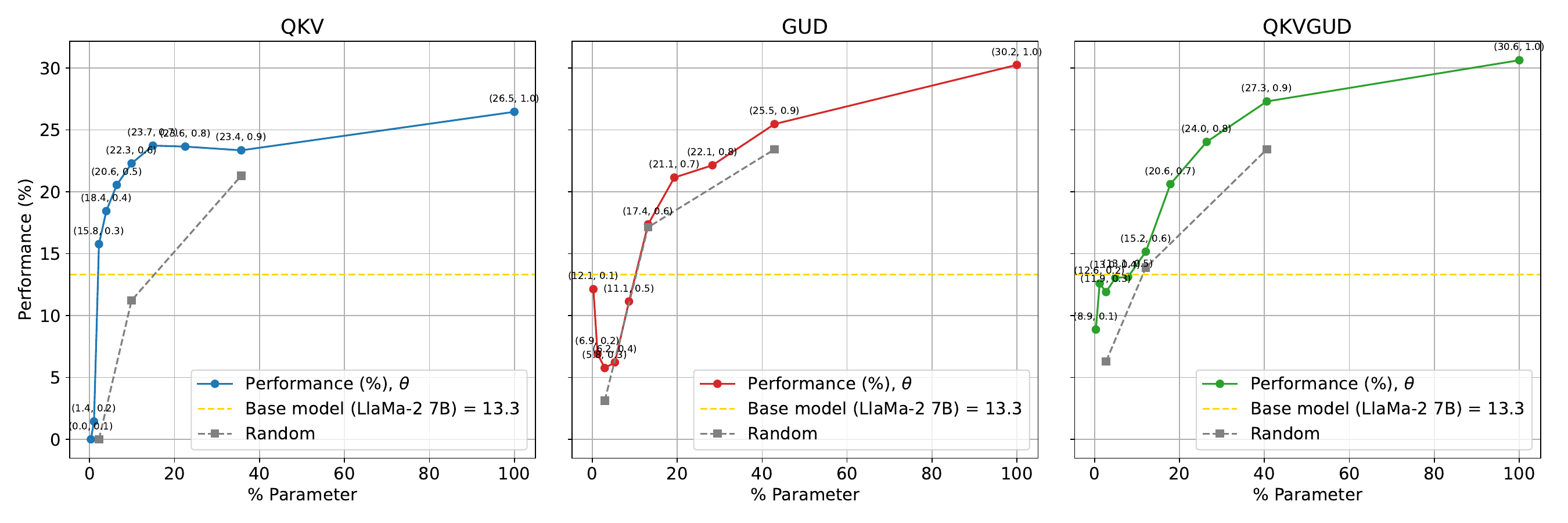} \hfill
  \caption {Performance on GSM8K versus the fraction of LoRA parameters activated (controlled by $\theta$), comparing QKV, GUD, and QKVGUD configurations. The dashed gray line represents random selection of the same fraction, and the horizontal yellow line is the base model’s (LLaMA2 7B) performance.}
    \vspace{0.5cm}
  \label{fig:theta_trend}
  
\end{figure*}

Table~\ref{tab:theta} summarizes our results on GSM8K, and Figure~\ref{fig:theta_trend} illustrates the overall trend. As $\theta$ increases, performance generally improves, but diminishing returns emerge near the high end. Notably, QKVGUD reaches close to its maximum accuracy even around $\theta$ between 0.8 and 0.9, activating only 30\%–40\% of LoRA parameters. In contrast, random selection of an equivalent fraction falls short, underscoring that \emph{targeting the most important} parameters is critical to strong performance.
In practice, $\theta$ can thus be tuned to strike a balance between \emph{absolute accuracy} and \emph{parameter budget}.

\begin{table}[t]
  \centering
  \begin{tabular}{lccc}
    \hline
    \textbf{\({\alpha}\)} & \textbf{\(\beta\)} & \textbf{Perf. (SFT)} & \textbf{Perf. (RL)} \\
    \hline
    0   & 0   & 18.88 & 19.03 \\
    0   & 0.5 & 18.12 & 25.25 \\
    0   & 1   & 18.35 & 23.58 \\
    0.5 & 0   & 23.65 & 25.85 \\
    0.5 & 0.5 & 24.79 & 29.57 \\
    0.5 & 1   & 24.94 & 31.01 \\
    1   & 0   & 27.75 & 27.75 \\
    1   & 0.5 & 26.99 & 31.92 \\
    1   & 1   & 27.14 & \textbf{34.37} \\
    \hline
  \end{tabular}
  \caption{
  \textbf{Influence of \(\alpha\) and \(\beta\) on two-stage fine-tuning (SFT then RL).}
All experiments use the GSM8K dataset with \emph{role-playing + voting} classification, 
\emph{SFT+RL} training, $\theta=0.9$, and the \texttt{QKVGUD} LoRA module.
``Perf. (SFT)'' measures model accuracy after the first stage,
while ``Perf. (RL)'' is the final accuracy after the second stage.
When $\alpha=\beta=1$, yielding the highest final score (in bold).
  }
  \vspace{-0.3cm}
  \label{tab:ab}
\end{table}

\begin{table*}[t]
\centering
\small
\begin{tabular}{c|ccc|ccc|ccc}
\hline
& \multicolumn{3}{c|}{\textbf{QKV}} 
& \multicolumn{3}{c|}{\textbf{GUD}} 
& \multicolumn{3}{c}{\textbf{QKVGUD}} \\
\hline
\textbf{$\theta$} 
& \textbf{\%Param} & \textbf{Perf.} & \textbf{Rand.} 
& \textbf{\%Param} & \textbf{Perf.} & \textbf{Rand.}
& \textbf{\%Param} & \textbf{Perf.} & \textbf{Rand.} \\
\hline
0 (Base LLaMA2 7B) & - & 13.3 & - & - & 13.3 & - & - & 13.3 & - \\
1 (Original LoRA) & 100 & 26.46 & - & 100 & 30.25 & - & 100 & 30.63 & - \\
\hline
0.1  & 0.30 & 0 & - & 0.22 & 12.13 & - & 0.32 & 8.87 & - \\
0.2  & 1.03 & 1.44 & - & 1.21 & 6.90 & - & 1.22 & 12.59 & - \\
0.3  & 2.20 & 15.77 & 0 & 2.89 & 5.76 & 3.11 & 2.72 & 11.90 & 6.29 \\
0.4  & 3.93 & 18.44 & - & 5.30 & 6.22 & - & 4.90 & 13.04 & - \\
0.5  & 6.39 & 20.55 & - & 8.61 & 11.14 & - & 7.91 & 13.12 & - \\
0.6  & 9.88 & 22.29 & 11.22 & 13.10 & 17.36 & 17.13 & 12.07 & 15.16 & 13.87 \\
0.7  & 14.90 & \textbf{23.73} & - & 19.30 & 21.15 & - & 17.87 & 20.62 & - \\
0.8  & 22.52 & 23.65 & - & 28.27 & 22.14 & - & 26.40 & 24.03 & - \\
0.9  & 35.70 & 23.35 & 21.30 & 42.90 & \textbf{25.47} & 23.43 & 40.56 & \textbf{27.3} & 23.43 \\
\hline
\end{tabular}
\caption{
\textbf{Performance on GSM8K with varying cutpoint (\(\theta\)) and different module configurations.}
For each \(\theta\), ``\%Param'' denotes the percentage of LoRA parameters activated,
``Perf.'' is the performance metric (accuracy), and ``Rand.'' is the performance 
when randomly selecting the same fraction of parameters.
All runs use LLaMA2 7B + LoRA with SFT.
}
\label{tab:theta}
\vspace{0.2cm}
\end{table*}

\begin{table*}[t]
\centering
\small
\renewcommand{\arraystretch}{1.1}  
\begin{tabular}{l l | c c | c c | c c | c c}
\hline
\multicolumn{2}{c|}{} 
& \multicolumn{2}{c|}{\textbf{GSM8K}} 
& \multicolumn{2}{c|}{\textbf{MMLU(Dolly)}} 
& \multicolumn{2}{c|}{\textbf{MMLU(Platypus)}} 
& \multicolumn{2}{c}{\textbf{HumanEval}} \\
\cline{3-10}
 & \textbf{Method} 
& \textbf{1 epoch} & \textbf{2 epoch} 
& \textbf{1 epoch} & \textbf{2 epoch}
& \textbf{1 epoch} & \textbf{2 epoch}
& \textbf{1 epoch} & \textbf{2 epoch} \\
\hline
\multirow{1}{*}{} 
& \textbf{\textbf{LLaMa2\,7B}} 
  & 13.3  & -- 
  & 46.48 & -- 
  & 46.48 & -- 
  & 14.39 & -- \\
\hline
\multirow{4}{*}{\textbf{LoRA}}
& LoRA \citeyearpar{Hu2021LoRALA}
  & 27.8  & 31.86 
  & 45.85 & 44.99 
  & 43.36 & 43.16 
  & 15.37 & 18.54 \\

& OLoRA \citeyearpar{Bykakyz2024OLoRAOL}
  & 29.57 & 32.83
  & 45.08 & 45.06
  & \textbf{45.16} & 45.26
  & 15.85 & 19.02 \\

& PiSSA \citeyearpar{Meng2024PiSSAPS}
  & 30.78 & 33.59
  & 22.95 & 23.27
  & 24.14 & 23.89
  & 21.95 & 24.39 \\

& PiSSA+RL
  & 30.63 & 37.45
  & 22.96 & 23.45
  & 23.98 & 23.92
  & 21.34 & 25.61 \\
\hline
\multirow{3}{*}{\textbf{Proposal}}
& *PiSSA($\theta=0.9$)  
  & 28.73 & 38.97
  & 23.10 & 24.07
  & 24.11 & 25.01
  & 22.56 & 26.22 \\

& *PiSSA($\theta=0.95$) 
  & \textbf{30.86} & \textbf{41.85}
  & 23.33 & 24.14
  & 24.41 & 25.38
  & \textbf{23.17} & \textbf{27.43} \\

& *LoRA($\theta=0.9$)
  & 27.07 & 34.57
  & \textbf{46.34} & \textbf{47.09}
  & 44.15 & \textbf{45.66}
  & 16.46 & 19.51 \\
\hline
\end{tabular}
\caption{\textbf{Comparison of LoRA‐based baselines versus our proposals on four benchmarks.} We use LLaMA2 7B as the base model; One or two training epochs are performed, as our proposal, PiSSA+RL specifically doing one SFT followed by one RL epoch. Our proposed methods (marked ) combine role‐play+voting data splitting and importance‐based parameter activation ($\theta$) with fully active shared region ($\alpha=\beta=1$).}
\label{tab:compare-final}
\end{table*}

\subsection{Utilizing Shared Parameters via $\alpha$ \& $\beta$}
Recall that $\alpha$ and $\beta$ (introduced in \S\ref{sec:two_stage}) control how many of the \emph{shared} parameters remain active in the SFT (System 1) and RL (System 2) stages, respectively. We fix $\theta=0.9$ for each system, then vary $\alpha$ and $\beta$ to measure their impact on training dynamics.
Table~\ref{tab:ab} shows results on GSM8K using QKVGUD LoRA. The “Performance (SFT)” column reflects accuracy after the first stage, while “Performance (RL)” is the final accuracy after the second stage. When both $\alpha = \beta = 1$, shared parameters remain fully active in both stages—maximizing the SFT checkpoint and yielding the best final score (34.37). Lower values of $\alpha$ or $\beta$ reduce the overlap, limiting early gains in SFT or hampering the RL stage’s capacity for multi‐step reasoning. In effect, a strong SFT foundation provides a “warm start” for RL (System 2), letting the model build deeper logic on top of its fast‐thinking skills.

\subsection{Final Performance and Baseline Comparisons}

We conclude by evaluating our approach alongside several LoRA‐based baselines (LoRA \citep{Hu2021LoRALA}, OLoRA \citep{Bykakyz2024OLoRAOL}, PiSSA \citep{Meng2024PiSSAPS}, PiSSA+RL) on four tasks: GSM8K, MMLU (trained with Dolly15K or Platypus), and HumanEval. Each column in Table \ref{tab:compare-final} corresponds to one of these tasks, trained for either one or two epochs. For instance, GSM8K is trained on its own data, whereas MMLU(Dolly) and MMLU(Platypus) use Dolly15K and OpenPlatypus, respectively. HumanEval relies on code-focused data (CodeAlpaca \citep{codealpaca}, CodeFeedback). The base model (LLaMA2 7B) is shown for reference, with no fine‐tuning.
Baselines typically perform two rounds of SFT, except for PiSSA+RL, which does one SFT epoch and then a RL epoch. Our proposals (marked with *) apply $\theta$-based parameter selection ($\theta=0.9$ or $\theta=0.95$) along with role‐play+voting data splits and fully active overlaps ($\alpha=\beta=1$) in the two‐stage training. 
Notably, PiSSA($\theta=0.9$ or $\theta=0.95$) uses about 40\% of the full LoRA parameters yet outperforms vanilla PiSSA, indicating that focusing on high‐importance subregions can yield stronger results.
Overall, our method achieves the best accuracy on GSM8K (41.85\%), surpassing PiSSA by about 12\% while using significantly fewer parameters. On MMLU, we also observe gains over standard LoRA and PiSSA, confirming that selectively activating only the most relevant parameters is both efficient and effective. Beyond PiSSA, our QKVGUD configuration activates only about 40\% of LoRA parameters (based on $\theta=0.9$ or $0.95$) for each system, yet still outperforms both full LoRA and random subsets of comparable size. As illustrated by the scatter plots in Figure~\ref{fig:weight_importance_plots}, these “top‐ranked” parameters form a highly specialized subregion for fast \& intuitive (System 1) vs.\ multi‐step (System 2) tasks. In other words, by focusing updates on those parameters that are important to each reasoning style, we realize the dual‐system analogy—distinct parameter subsets excel at quick, intuitive SFT or stepwise RL—while lowering active parameter usage.

\section{Limitations}
Our experiments confirm that selectively activating LoRA parameters for System 1 and System 2 yields clear benefits in both performance and parameter efficiency. By combining role‐play‐based data splitting with importance‐driven parameter partitioning, we effectively approximate the dual‐process paradigm within an LLM. Nonetheless, there are a few limitations:
\textbf{(1) Multi‐Model Annotations.} Although using multiple teacher LLMs improves labeling quality, it adds computational overhead and presupposes access to diverse, high‐capacity models.
\textbf{(2) Granularity of Task Partitioning.} Our approach treats tasks at a coarse level (System 1 vs.\ System 2). More nuanced distinctions (e.g., intermediate steps or partial multi‐hop reasoning) may require finer‐grained analysis.
\textbf{(3) Applicability to Other Architectures.} We have shown results on LLaMA2 7B; generalizing to other model families (e.g., decoder‐encoder hybrids) may require adjustments in how LoRA parameters are attached and scored.

\section{Conclusion}
We presented a dual‐system PEFT framework inspired by Thinking, Fast and Slow, wherein System 1 and System 2 “subregions” of LoRA parameters address fast, intuitive tasks vs.\ slower, multi‐step reasoning. Our pipeline (i) classifies queries via multi‐model role‐playing and voting, (ii) determines each LoRA parameter’s importance relative to System 1 or System 2, and (iii) fine‐tunes in two stages—SFT for intuitive response, then RL for deeper logic.
Across GSM8K, MMLU, and HumanEval, we find that focusing updates on top‐ranked parameters not only cuts active parameter usage (often to 40\% or less) but also surpasses baseline PEFT methods that uniformly fine‐tune larger parameter sets. By assigning each subregion to a distinct “cognitive” mode, we effectively reconcile fast vs.\ slow thinking within a single LLM. We believe this “subregion specialization” opens new directions for cognitively guided LLM adaptations, enabling more efficient models that still excel at both intuitive and methodical reasoning.
\bibliography{custom}

\appendix
\section*{Content of Appendix}
\begin{itemize}[noitemsep, topsep=0pt]
    \item[\ref{sec:Implemetation Details}] Implemetation Details
    \begin{itemize}
        \item[\ref{sec:Data Preparation and System 1/2 Labeling}] Data Preparation and System 1/2 Labeling
        \item[\ref{sec:LoRA Integration}] LoRA Integration
        \item[\ref{sec:Importance Scoring and Parameter Activation}] Importance Scoring and Parameter Activation
        \item[\ref{sec:Supervised Fine-Tuning (System 1)}] Supervised Fine-Tuning (System 1)
        \item[\ref{sec:Reinforcement Learning (System 2)}] Reinforcement Learning (System 2)
        \item[\ref{sec:Training Footprint}] Training Footprint
    \end{itemize}
    \item[\ref{sec:Additional Experiments}] Additional Experiments
    \begin{itemize}
        \item[\ref{sec:Generalization to a Different Backbone}] Generalization to a Different Backbone (Qwen-2.5-7B on GSM8K)
        \item[\ref{sec:Broader PEFT Baselines on LLaMA-2-7B}] Broader PEFT Baselines on LLaMA-2-7B
        \item[\ref{sec:Order-of-Training Ablation}] Order-of-Training Ablation: SFT→RL vs. RL→SFT
    \end{itemize}
\end{itemize}
\vspace{-0.3cm}

\section{Implemetation Details}
\label{sec:Implemetation Details}

Here we summarize the implementation of LoRA-PAR corresponding to Section \ref{sec:Method}. We first construct a reusable S1/S2–labeled pool via role-play + voting, then estimate element-wise LoRA importance on small S1/S2 subsamples and activate parameter subregions by a cumulative-importance threshold $\theta$, with shared elements governed by $\alpha$ and $\beta$. Stage 1 performs SFT on the S1-activated subregion; Stage 2 applies GRPO-based RL on the S2-activated subregion with a composite reward that prioritizes answer correctness while stabilizing output format. Element-wise gradient hooks enforce the masks throughout, all non-LoRA weights remain frozen, and we report hardware/software settings and training footprint to support reproducibility. \footnote{The code of our proposal can be viewed at \href{https://github.com/EmergencerOnEarth/LoRA-PAR}{https://github.com/EmergencerOnEarth/LoRA-PAR}} 
\vspace{-0.1cm}
\subsection{Data Preparation and System 1/2 Labeling}
\label{sec:Data Preparation and System 1/2 Labeling}
Following Section \ref{sec:Method}, we first split data into System 1 (S1) vs. System 2 (S2) using role-play + voting. In practice we produce a single labeled pool and reuse it across experiments. For importance calculation we subsample only 100 S1 and 100 S2 samples respectively to bound cost while keeping a stable signal from each class. During tokenization, we concatenate the prompt and the target; labels for prompt tokens are set to -100, so the loss is computed only on the answer segment.
\vspace{-0.1cm}
\subsection{LoRA Integration}
\label{sec:LoRA Integration}
LoRA adapters to the MLP projections \texttt{gate\_proj}, \texttt{up\_proj}, and \texttt{down\_proj} (GUD)  are attached for the S1 stage; other attachment patterns (e.g., QKV) are compatible with the same pipeline. Unless noted otherwise, we use rank r=8, \texttt{lora\_alpha}=64, and set LoRA dropout to 0.05 for training (initialized as 0.0 then updated programmatically). All non-LoRA parameters are frozen (\texttt{requires\_grad}=False). To avoid all importance scores are set to zero, we explicitly initialize LoRA A/B matrices with respect to $\mathcal{N}(0, 0.01)$.
\vspace{-0.1cm}
\subsection{Importance Scoring and Parameter Activation}
\label{sec:Importance Scoring and Parameter Activation}
To implement the parameter "subregion" idea from Section \ref{sec:Method}, we estimate element-wise importance for all LoRA weights using a masked-loss, second-order proxy computed over the 100-sample S1 and S2 mini-corpora. For each batch we back propagation once and accumulate, per LoRA tensor element, (i) the sum of \texttt{$grad * weight$} and (ii) the sum of \texttt{$(grad * weight)^2$}. We then apply L2-normalize to the importance and choose a cut-point so that the cumulative sum reaches a threshold $\theta$; all elements with normalized score greater than cut-point are marked "activated." Before SFT we register per-element gradient hooks that zero out gradients where the mask is 0, implementing element-wise freezing inside each LoRA tensor (rather than whole-module freezing). We report the fraction of trainable elements with a simple counter to verify the effective active-parameter rate.
\vspace{-0.1cm}
\subsection{Supervised Fine-Tuning (System 1)}
\label{sec:Supervised Fine-Tuning (System 1)}
SFT trains only the S1-activated subregion plus any globally unfrozen LoRA elements. We use TRL's SFTTrainer with: max sequence length 512, learning rate 2e-4, cosine schedule, warm-up ratio 0.03, per-device batch size 1 with gradient accumulation 32, and 1 epoch. After SFT all element-wise freeze hooks are removed so that the subsequent stage (RL for S2) can re-apply its own mask. 
\vspace{-0.1cm}
\subsection{Reinforcement Learning (System 2)}
\label{sec:Reinforcement Learning (System 2)}
For the RL phase we re-partition the LoRA subspace to target System-2 reasoning. Using the S1 SFT checkpoint as initialization, we activate all elements in $\omega_{2-only}$ together with the top $\beta$ fraction of the shared region ranked by S2 importance. Activation is enforced at element granularity via gradient hooks that zero the per-element gradient wherever the mask is zero, so only the intended LoRA subregion is trainable while all base weights remain frozen. We optimize with TRL's GRPO package; the trainer maintains the frozen SFT policy internally for KL control. Unless stated otherwise, training uses a learning rate of 5e-6 with Adam, weight decay 0.1, cosine decay with warm-up ratio 0.1, group size 8 samples per prompt, maximum prompt and completion lengths of 128 and 256 tokens respectively, 1 epoch, and gradient-norm clipping at 0.1; checkpoints are saved periodically and all element-wise hooks are removed after training to allow subsequent reconfiguration if needed.

Rewards are shaped to make correctness the dominant learning signal while stabilizing structure. Each generated sample receives the sum of five components: a small dense XML well formatted score that rewards the presence and proper closure of tags with mild penalties for trailing content; a permissive format check and a stricter template match that together encourage consistent scaffolding; a type check that grants a bonus if the extracted answer is an integer; and a correctness bonus that yields 2.0 for exact match between the extracted answer and the gold answer. The first four terms are each scaled to 0.5, whereas the 2.0 correctness term provides the main learning pressure toward solving the task. 
\vspace{-0.1cm}
\subsection{Training Footprint}
\label{sec:Training Footprint}
Runs use a single NVIDIA L20 (48 GB). On GSM8K, SFT completes in under one hour per epoch; the RL/GRPO epoch is longer ($\approx$ 13 h). Because we freeze low-importance LoRA elements, peak training memory and optimizer state size scale with the active subregion; inference latency/memory remain comparable to a standard LoRA model.
\vspace{-0.2cm}
\section{Additional Experiments}
\label{sec:Additional Experiments}
\vspace{-0.1cm}
\subsection{Generalization to a Different Backbone (Qwen-2.5-7B on GSM8K)}
\label{sec:Generalization to a Different Backbone}
To assess whether LoRA-PAR is tied to a specific model family, we repeat the main pipeline on Qwen-2.5-7B under the same recipe as in the proposal, unless otherwise noted: LoRA on Q/K/V/Gate/Up/Down, rank = 8; importance threshold $\theta=0.9$; shared-subset activation $\alpha=\beta=1$; one epoch of SFT followed by one epoch of RL (GRPO). For data labeling we used DeepSeek-R1 role-playing as Qwen-2.5-7B (single model, no voting) to obtain the system 1/2 labels. Table \ref{tab:qwen_single} shows that our method achieves 68.69\% after SFT and 81.65\% after RL, outperforming strong LoRA-family baselines and also exceeding the base model's 1-shot score. This indicates that the our propoal is not specific to LLaMA-2.

\begin{table}[t]
  \centering
  \begin{tabular}{lcc}
    \hline
    \textbf{Method} & \textbf{1 epoch} & \textbf{2 epoch} \\
    \hline
    \multicolumn{3}{l}{\emph{Base model}} \\
    Qwen 2.5 7B (0-shot) & 21.91 & -- \\
    Qwen 2.5 7B (1-shot) & 62.02 & -- \\
    \hline
    \multicolumn{3}{l}{\emph{LoRA family}} \\
    LoRA   & 67.32 & 68.61 \\
    OLoRA  & 65.43 & 67.32 \\
    PiSSA  & 55.80 & 57.24 \\
    LoRA+RL & 67.25 & 69.83 \\
    \hline
    \multicolumn{3}{l}{\emph{Proposal}} \\
    \textbf{LoRA-PAR}  & \textbf{68.69} & \textbf{81.65} \\
    \hline
  \end{tabular}
  \caption{\textbf{GSM8K on Qwen-2.5-7B.} Baselines use two SFT epochs (except LoRA+RL / LoRA-PAR where the second epoch is RL). The LoRA-PAR is with $\theta{=}0.9,\ \alpha{=}\beta{=}1$.}
  \label{tab:qwen_single}
\end{table}

\begin{table}[t]
  \centering
  \begin{tabular}{lcc}
    \hline
    \textbf{Method / Schedule} & \textbf{1 epoch} & \textbf{2 epoch} \\
    \hline
    \multicolumn{3}{l}{\emph{Base model}} \\
    Qwen 2.5 7B (0-shot) & 21.91 & -- \\
    Qwen 2.5 7B (1-shot) & 62.02 & -- \\
    \hline
    \multicolumn{3}{l}{\emph{Two-stage schedules}} \\
    \textbf{SFT $\rightarrow$ RL (proposal)} & 68.69 & \textbf{81.65} \\
    \textbf{RL $\rightarrow$ SFT}            & \textbf{70.81} & 68.46 \\
    \hline
  \end{tabular}
  \caption{\textbf{Order-of-training ablation on Qwen-2.5-7B (GSM8K).} For two-stage schedules, ``1 epoch'' is phase~1 and ``2 epoch'' is phase~2.}
  \label{tab:order_single}
\end{table}

\begin{table}[t]
  \begin{tabular}{lcc}
    \hline
    \textbf{Method} & \textbf{1 epoch} & \textbf{2 epoch} \\
    \hline
    \multicolumn{3}{l}{\emph{Base model}} \\
    LLaMA-2-7B & 13.3 & -- \\
    \hline
    \multicolumn{3}{l}{\emph{LoRA family}} \\
    LoRA   & 27.8  & 31.86 \\
    OLoRA  & 29.57 & 32.83 \\
    PiSSA  & 30.87 & 33.59 \\
    PiSSA+RL & 30.63 & 37.45 \\
    \hline
    \multicolumn{3}{l}{\emph{Other PEFT}} \\
    Prompt Tuning & 22.37 & 26.31 \\
    P-Tuning      & 24.18 & 26.76 \\
    Prefix-Tuning & 25.7  & 30.02 \\
    IA3           & 22.97 & 26.46 \\
    \hline
    \multicolumn{3}{l}{\emph{Proposal}} \\
    \textbf{PiSSA} ($\theta{=}0.9$)  & 28.73 & 38.97 \\
    \textbf{PiSSA} ($\theta{=}0.95$) & \textbf{30.86} & \textbf{41.85} \\
    \textbf{LoRA}  ($\theta{=}0.9$)  & 27.07 & 34.57 \\
    \hline
  \end{tabular}
  \caption{\textbf{GSM8K on LLaMA-2-7B with broader PEFT baselines.} Non-LoRA baselines are trained with two SFT epochs; proposal rows use importance-based activation with the indicated $\theta$.}
  \label{tab:llama_peft_single}
\end{table}

\subsection{Broader PEFT Baselines on LLaMA-2-7B}
\label{sec:Broader PEFT Baselines on LLaMA-2-7B}
We additionally compared against non-LoRA PEFT variants on LLaMA-2-7B, training each baseline for two SFT epochs. As reported in Table \ref{tab:llama_peft_single}, these alternatives (prompt-tuning, P-tuning, prefix-tuning, IA3) underperform the LoRA family under our setup. Our proposal, when applied on top of PiSSA with $\theta=0.95$, reaches 41.85\%, exceeding all compared PEFT baselines.

\subsection{Order-of-Training Ablation: SFT→RL vs. RL→SFT}
\label{sec:Order-of-Training Ablation}
To probe the training-order choice, we ran a reverse schedule on Qwen-2.5-7B / GSM8K using the same LoRA configuration as in \ref{sec:Generalization to a Different Backbone}. Table \ref{tab:order_single} shows that starting with RL yields 70.81\% after the first epoch, slightly above SFT-first (68.69\%). However, the subsequent SFT epoch reduces performance to 68.46\%, whereas our original SFT→RL pipeline continues to improve to 81.65\%. A plausible interpretation is that RL initially cultivates multi-step reasoning behaviors, which next-token SFT may overwrite; by contrast, SFT establishes a strong imitational base that RL can refine without eroding emergent reasoning skills.

\end{document}